\documentclass[letterpaper, 10 pt, conference]{ieeeconf}
\IEEEoverridecommandlockouts    % This command is only needed if
\usepackage[textwidth=4cm,disable]{todonotes}

\usepackage{graphics}           
\usepackage{times}              
\usepackage{amsmath}            
\usepackage{amssymb}            
\usepackage{graphicx}
\usepackage{algorithm}
\usepackage[noend]{algpseudocode}
\usepackage{booktabs}
\usepackage{color}

\definecolor{instructioncolor}{rgb}{.5,.5,.5}

\usepackage[font=small]{caption}

\def\secref#1{Sec.~\ref{#1}}
\def\figref#1{Fig.~\ref{#1}}
\def\tabref#1{Tab.~\ref{#1}}
\def\eqref#1{Eq.~(\ref{#1})}

\def\ie{\textit{i.e.}}

\def\eg{\textit{e.g.}}

\def\signbbox{B}
\def\signplace{p}

\def\signdir{d}

\def\signpossdirs{\mathcal{D}}
\def\signlabel{l}

\def\setpred{\mathcal{P}}
\def\setgt{\mathcal{G}}

\def\setbbox{\mathcal{B}}
\def\setcues{\mathcal{C}}

\makeatletter
\usepackage{xspace}
\DeclareRobustCommand\onedot{\futurelet\@let@token\@onedot}
\def\@onedot{\ifx\@let@token.\else.\null\fi\xspace}
\def\eg{e.g\onedot} 
\def\ie{i.e\onedot} 
 
\def\etc{etc\onedot} 
 
\def\etal{{et al}\onedot}
\def\etalcite#1{\etal~\cite{#1}}

\makeatother

\usepackage{array}
\newcolumntype{L}[1]{>{\raggedright\let\newline\\\arraybackslash\hspace{0pt}}m{#1}}
\newcolumntype{C}[1]{>{\centering\let\newline\\\arraybackslash\hspace{0pt}}m{#1}}
\newcolumntype{R}[1]{>{\raggedleft\let\newline\\\arraybackslash\hspace{0pt}}m{#1}}

\let\labelindent\relax
\usepackage{siunitx}
\usepackage{enumitem}
\usepackage{subcaption}
\usepackage{booktabs}
\usepackage[table]{xcolor}
\usepackage{arydshln}
\usepackage{hyperref}
\title{\LARGE \bf Sign Language: Towards Sign Understanding for Robot Autonomy}

 \author{Ayush Agrawal \ddag $\ast$ \and Joel Loo\dag $\ast$ \and  Nicky Zimmerman\dag $\ast$ \and David Hsu\dag % <-this % stops a space
\thanks{\dag Smart Systems Institute, National University of Singapore.} \thanks{\ddag Northeastern University}  \thanks{ $\ast$ Equal contribution.} }

\begin{document}   
\maketitle
\thispagestyle{empty}
\pagestyle{empty}

%%%%%%%%%%%%%%%%%%%%%%%%%%%%%%%%%%%%%%%%%%%%%%%%%%%%%%%%%%%%%%%%%%%%%%%%%%%%%%%%
\begin{abstract}

%\textit{Navigational signs} are ubiquitous in human environments to aid scene understanding and navigation, but are underutilized by robot systems. We argue that they offer a direct source of top-down scene understanding, explicitly encoding human-aligned semantic entities and spatial relations that robots would otherwise need to infer from raw sensor data. Interpreting such signs has been challenging due to open-vocabulary and reasoning requirements, but advances in Vision-Language Models (VLMs) make this feasible. We introduce the task of \textit{navigational sign understanding} which parses locations and associated directions from signs. We posit that sign understanding can enable robust scene understanding and hence diverse downstream applications from localization to semantic navigation. To this end, we curate a benchmark for this task: we propose appropriate evaluation metrics, and offer a test set capturing signs with varying complexity and design across a range of public spaces, from hospitals to shopping malls to transport hubs. We also provide a baseline approach using VLMs, and demonstrate their promise on navigational sign understanding. Code and dataset will be released.

\textit{Navigational signs} are common aids for human wayfinding and scene understanding, but are underutilized by robots. We argue that they benefit robot navigation and scene understanding, by directly encoding privileged information on actions, spatial regions, and relations. Interpreting signs in open-world settings remains a challenge owing to the complexity of scenes and signs, but recent advances in vision-language models (VLMs) make this feasible. To advance progress in this area, we introduce the task of \textit{navigational sign understanding} which parses locations and associated directions from signs. We offer a benchmark for this task, proposing appropriate evaluation metrics and curating a test set capturing signs with varying complexity and design across diverse public spaces, from hospitals to shopping malls to transport hubs. We also provide a baseline approach using VLMs, and demonstrate their promise on navigational sign understanding. Code and dataset are available on Github 
\footnote{\url{https://github.com/AdaCompNUS/Sign-Understanding}}.

\end{abstract}

%%%%%%%%%%%%%%%%%%%%%%%%%%%%%%%%%%%%%%%%%%%%%%%%%%%%%%%%%%%%%%%%%%%%%%%%%%%%%%%%
\section{Introduction} \label{sec:intro}

%%%%%%%%%%%%%%%%%%%
%% WHY: 
% First, answer the WHY question: Why is that relevant? Why should I be
% motivated to read the paper? Why should I care? (1 paragraph, 2-5 sentences)

%% Scene understanding context
%% Throw 3 main arguments out
%% Human motivation
%% Navigational signs are under-utilized

% Scene understanding is an essential component in enabling autonomous agents, contributing to the agent's ability to localize, navigate and plan. Previous research works focused on identifying objects and reasoning about their spatial relationships by extracting geometric and semantic features~\cite{peng2023cvpr, gu2024icra}. The importance of semantic scene understanding is inspired by humans, who rely on symbols to communicate~\cite{grouchy2016sr}, use semantics to reason and extract symbolic information from their environment. Signs are a rich source of symbolic information utilized by humans, particularly navigational signs that are placed to aid for wayfinding. However, leveraging signs for scene understanding has not been explored yet, despite the fact that comprehensive scene understanding extends also to utilizing textual and symbolic information to speculate about unseen parts of the environment. Navigational signs allow us to extend our field of view beyond what can be captured with sensors locally, by exploiting relevant and intentional information about the environment, expressed directly in a level of abstraction that commonly requires intensive computation to infer. 

\textbf{Navigational signs} are ubiquitous elements of human environments designed to help humans navigate, and to aid spatial awareness and understanding in large, complex scenes~\cite{gibson2009pap, wang2023thesis}. They typically use text and symbols to denote \textit{locations} at different levels of abstraction (\eg{} rooms, floors, buildings) and associate them with \textit{directional} information. We argue that signs are rich, accessible sources of privileged information for robots. Signs facilitate \textit{navigation} by conveying local, granular actions for reaching goals specified in natural language. As sources of discrete, symbolic information, signs align well with the growing shift toward topo-semantic navigation~\cite{garg2024robohop}. Signs provide information for rich, human-aligned \textit{scene understanding}---a task that extends beyond scene mapping to capture abstractions (\eg{}, objects, regions) and relations among them~\cite{mascaro2024scenerepreview}. Existing approaches to scene understanding are largely bottom-up: spatial abstractions and relations are inferred from direct sensor observations of the physical space. In contrast, signs enable a top-down process: they directly communicate human-specified spatial abstractions and relations, including for unseen regions and entities out of the robot's line-of-sight.

Despite their potential, navigational signs remain underutilized in robotics. Prior work has mainly focused on classifying fixed sets of sign categories, or on exploiting textual information in signs---as landmarks for localization~\cite{cui2021iros, zimmerman2022iros} or to annotate locations in semantic mapping~\cite{case2011icra}. We identify two key challenges limiting the use of navigational signs in the open world. First, parsing signs poses a challenge since they contain open-vocabulary text and open-set symbols, requiring semantic reasoning to interpret associations between elements of a sign (\eg{} locations, directions). Second, detecting signs can be difficult as scenes contain diverse distractors which resemble but are semantically distinct from signs (\eg{} displays, advertisements). Recent advances in vision-language models (VLMs)~\cite{zhang2024tpami, deitke2024arxiv} offer powerful open-world understanding and reasoning, providing a means to address both \textit{sign} and \textit{scene complexity}.

\begin{figure}[t]
  \centering
  \includegraphics[clip, width=0.95\linewidth]{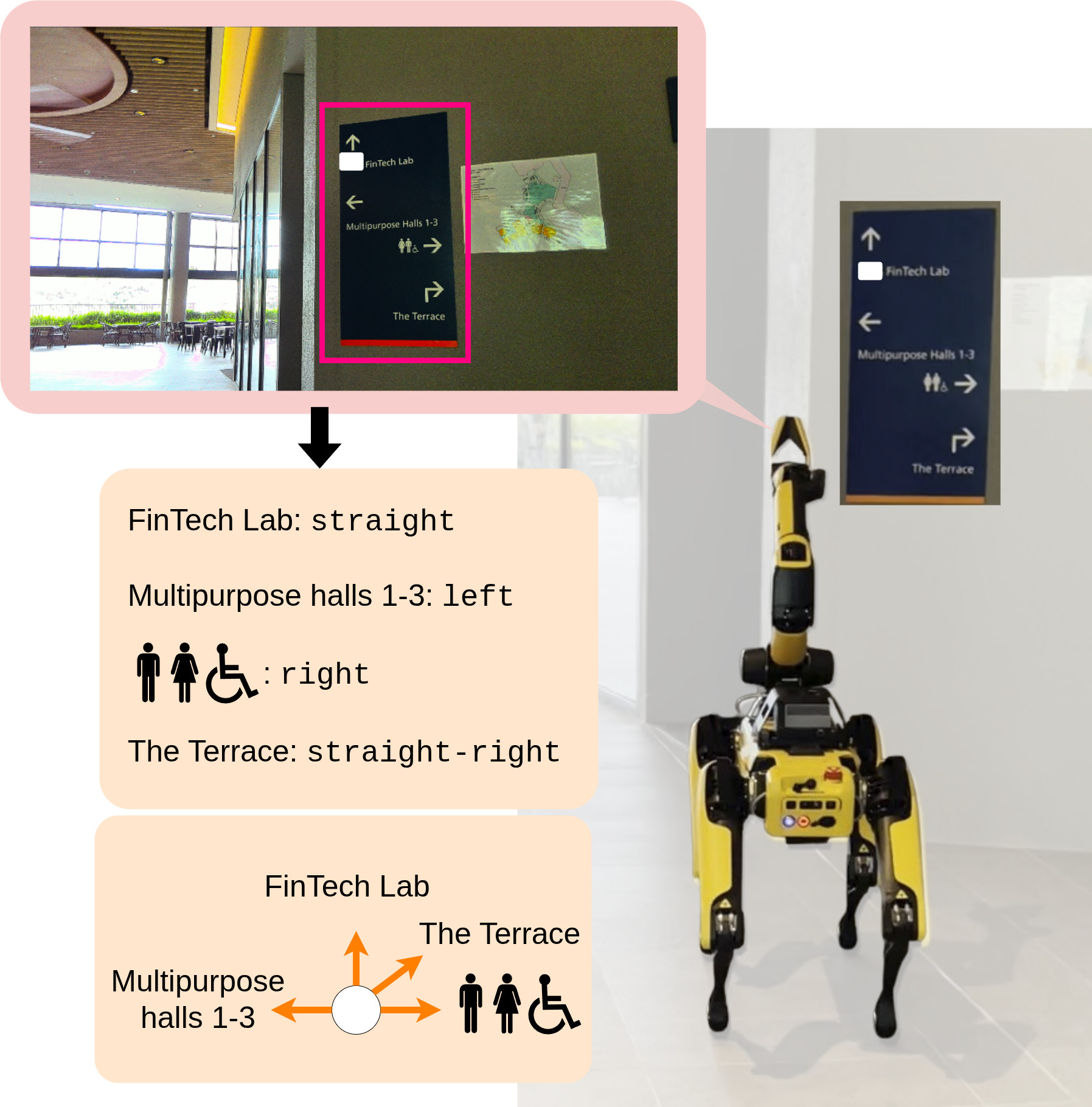}
  \caption{\textbf{Sign understanding in the wild with Spot robot.} Navigational signs are aids for navigation and scene understanding. We introduce the \textit{navigational sign understanding} task and design a baseline system to detect and parse signs on real hardware.}
  \label{fig:motivation} 
\end{figure}

\begin{figure*}[!t]
  \includegraphics[width=\linewidth]{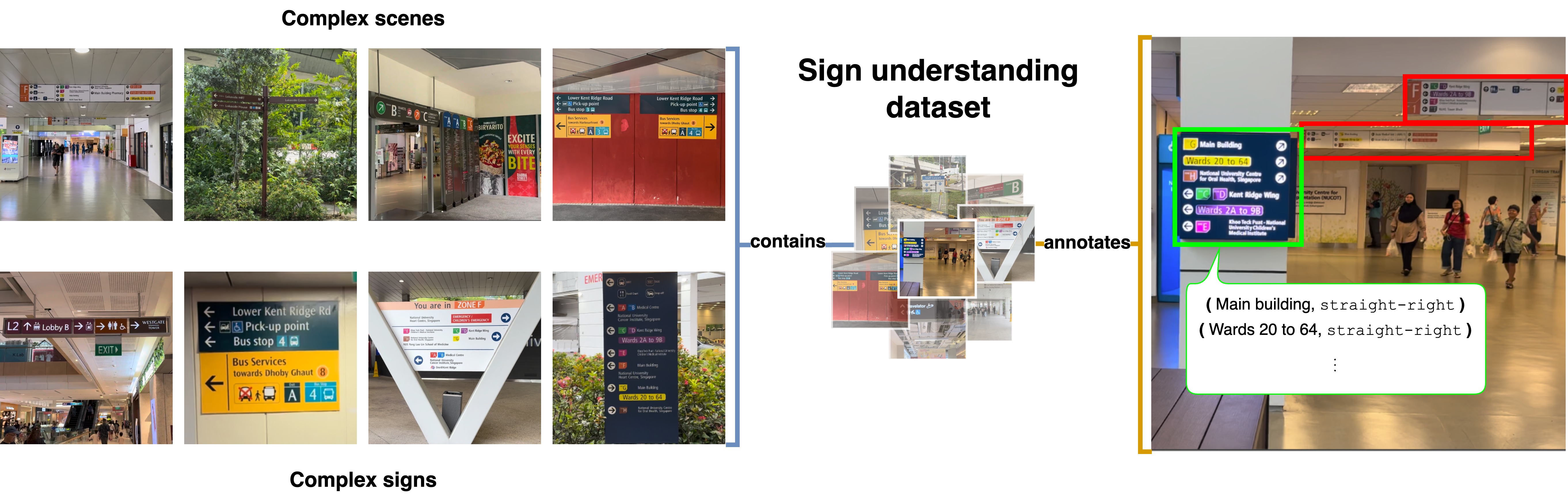}
  \caption{\textbf{Dataset to evaluate navigational sign understanding.} The dataset captures diverse signs across a varied range of scenes. These span across hospitals, transit hubs, malls, campuses, outdoor parks \etc{}, and include scenes with multiple signs, distractors and varying conditions, \eg{} illumination. Signs in the dataset reflect real-world complexity, including signs using diverse symbols, signs that are stylized or have unusual appearance, or signs that are information-dense. The dataset's human-annotated ground truth provides the bounding boxes for observed signs. Human readable signs are further annotated with the corresponding navigational cues.}
  \label{fig:dataset}
\end{figure*}

% The main contribution of this paper is to introduce the task of navigational sign understanding, and explore how it can enable \textit{top-down} scene understanding on real robots. We develop an open-source benchmark to evaluate performance on this task: we propose metrics and curate a test set of signs from diverse public spaces (\eg{}, hospitals, malls, subway stations). We also provide a baseline that uses VLMs to detect and interpret navigational signs. To spur downstream robotics applications for sign understanding, we release a plug-in ROS module that integrates sign understanding capabilities on real robots, and demonstrate it on a quadruped.

% The main contribution of this paper is to introduce the task of navigational sign understanding, and provide a proof-of-concept approach that can be deployed on real hardware, potentially enabling diverse downstream applications in navigation and scene understanding. We develop an open-source benchmark to evaluate performance on this task: we propose metrics and curate a test set of signs from diverse public spaces (\eg{}, hospitals, malls, subway stations). We design a baseline that uses VLMs to detect and interpret navigational signs. A key engineering challenge of sign understanding lies in selecting the appropriate viewpoint that enables a good view of the sign while aligning with its intended viewing direction. We implement a plug-in ROS module for real robot hardware that integrates sign understanding capabilities with active viewpoint selection, and demonstrate it on a quadruped.
The main contribution of this paper is to introduce the task of \textbf{navigational sign understanding} and present a proof-of-concept system deployable on real hardware, enabling downstream applications in robot navigation and scene understanding. To support this task, we release an open-source benchmark combining a proposed set of evaluation metrics with a curated test set of signs from diverse public spaces (\eg{} hospitals, malls, subway stations). We design a VLM-based baseline for detecting and interpreting signs. A key engineering challenge of this task is to select viewpoints that provide both clear visibility and alignment with the sign’s intended viewing direction. To address this, we implement a ROS module integrating sign understanding with active viewpoint selection and demonstrate it on a quadruped robot.

\section{Related Work} \label{sec:related}  

Previous research tackled the classification of indoor and traffic signs, as well as the recognition of text in multiple domains.  While navigational signs are commonplace and part of everyday life for humans, no prior work addresses the detection and parsing of navigational signs, and specifically the complexity of associating open-set vocabulary and directional cues.

\subsection{Traffic Sign Recognition }
The importance of traffic sign recognition~(TSR) for intelligent vehicles have been recognized in previous decades~\cite{de1997tie, %mogelmose2012tits, 
stallkamp2012nn}. Earlier approaches used hand-crafted features to detect and classify the sign~\cite{de2003ivc, 
%hoferlin2009ivs, 
xu2016sensors}, yielding subpar performance. The increased interest in autonomous vehicles, advancements in deep learning and the availability of larger datasets boosted the research in this domain, resulting in more robust performance~\cite{wang2023nca, zhang2022hccis}. The task of traffic sign recognition includes detecting and classifying traffic signs using a set of pre-defined categories. It does not extend to parsing compound navigational signs that include both open vocabulary text, symbols and directional arrows.

\subsection{Indoor Sign Detection}
The problem of indoor sign detection~(ISD) differs from TSR in two speacts. First, traffic signs are standardized while indoor signage is not. Second, the placement of traffic signs allows full visibility and the signs' appearance is easily segmented from the background, while indoor signs are often surrounded by clutter and might be partially occluded~\cite{almeida2019visi}. Similarly to TSR, the earlier classic approaches~\cite{kunene2016sfics, Wang2013nmahib} were surpassed by learning-based approaches~\cite{cheraghi2021jtpd, de2024elticom}. As in the case of TSR, ISD remained limited to the detection and classification of a close set of classes, lacking the ability to extract spatial relationships between places represented as symbols and texts. 

\subsection{Text Spotting}
Text spotting, which refers to detecting text in an image and then recognizing the characters, has also experienced a leap in performance with the introduction of neural network architectures~\cite{liao2020aaai, yang2022cvpr}.
%shi2015arxiv, smith2007icdar}. 
Recent models such as PaddleOCR~\cite{du2020arxiv} and TesseractOCR~\cite{smith2007icdar} reliably extract textual information from standardized signs, but this alone is not sufficient to facilitate navigational sign understanding. 

The existing works are unable to reason about the spatial relationship between the detected place indicators and the world. The current research on sign recognition is also limited to a closed set of classes. To the best of our knowledge, the sign understanding task was not formally formulated yet and no attempt at a principled solution was made. Under the task of sign understanding, we aim to bring together the detection and recognition of symbols and text, with open set labels and association between the detected place indicators and their relative location.  

%%%%%%%%%%%%%%%%%%%%%%%%%%%%%%%%%%%%%%%%%%%%%%%%%%%%%%%%%%%%%%%%%%%%%%%%%%%%%%%%
\section{Task} 
\label{sec:task}
% Navigational signs are a unifying language for wayfinding in public spaces, where each sign conveys unique information. Signs are classified into four categories~\cite{gibson2009pap}: (i) orientational signs, such as venue maps and floor plans, offer an overview of the environment, (ii) regulatory signs inform about the required or prohibited behavior, (iii) identification/locational signs provide information about a current location, (iv) directional signs  indicates the direction of travel to specific locations. 

We introduce the task of \textbf{navigational sign understanding}, which extracts \textit{navigational cues} from signs. Given an input RGB image, we seek to first \textit{detect} navigational signs, then to \textit{recognize} the navigational cues in each sign. A navigational cue is a specific location indicated on the sign, and the direction associated with it. Specifically, the two subtasks of navigational sign understanding are:

\textbf{Navigational sign detection:} To identify all navigational signs in a given RGB image input, $I$. The output is $\setbbox = \{\signbbox_1,\dots,\signbbox_N\}$, where $\signbbox_n$ is a 2D bounding box of a navigational sign in $I$. Specifically, navigational signs should be signs containing \textit{both} location and direction information, thus excluding signs only containing locational information like room identification signs on doors.

\textbf{Navigational sign recognition:} To extract the set of navigational cues in an input navigational sign. The input sign is represented as an image crop $I_n$. The set of navigational cues for a given sign indicating $M$ locations is:
\begin{equation}
    \setcues_n={(\signplace_1,\signlabel_1,\signdir_1),\dots,(\signplace_M,\signlabel_M,\signdir_M)}
\end{equation} 
While locations in a sign may be indicated in both text or symbol form, we express all locations as plaintext strings for ease of representation. $\signplace_m$ is a single unique location indicated on the sign, specified as a string. $\signlabel_m\in\{\text{text}, \text{symbol}\}$ identifies $\signplace_m$'s original form in the sign. $\signdir_m\in\signpossdirs$ is the indicated direction that is associated with $\signplace_m$. Across signs, directions tend to be expressed mainly with eight cardinal directions. We specify the set of possible directions, $\signpossdirs$, as $\{\texttt{None}, \texttt{straight}, \texttt{right},\dots,\texttt{straight-left}\}$, where $\texttt{None}$ denotes purely locational cues not associated to any direction.

Navigational sign understanding combines detection and recognition. Detection results provide image crops of each identified sign, and recognition further extracts the set of cues from each crop.

\begin{figure*}[!t]
  \includegraphics[width=\linewidth]{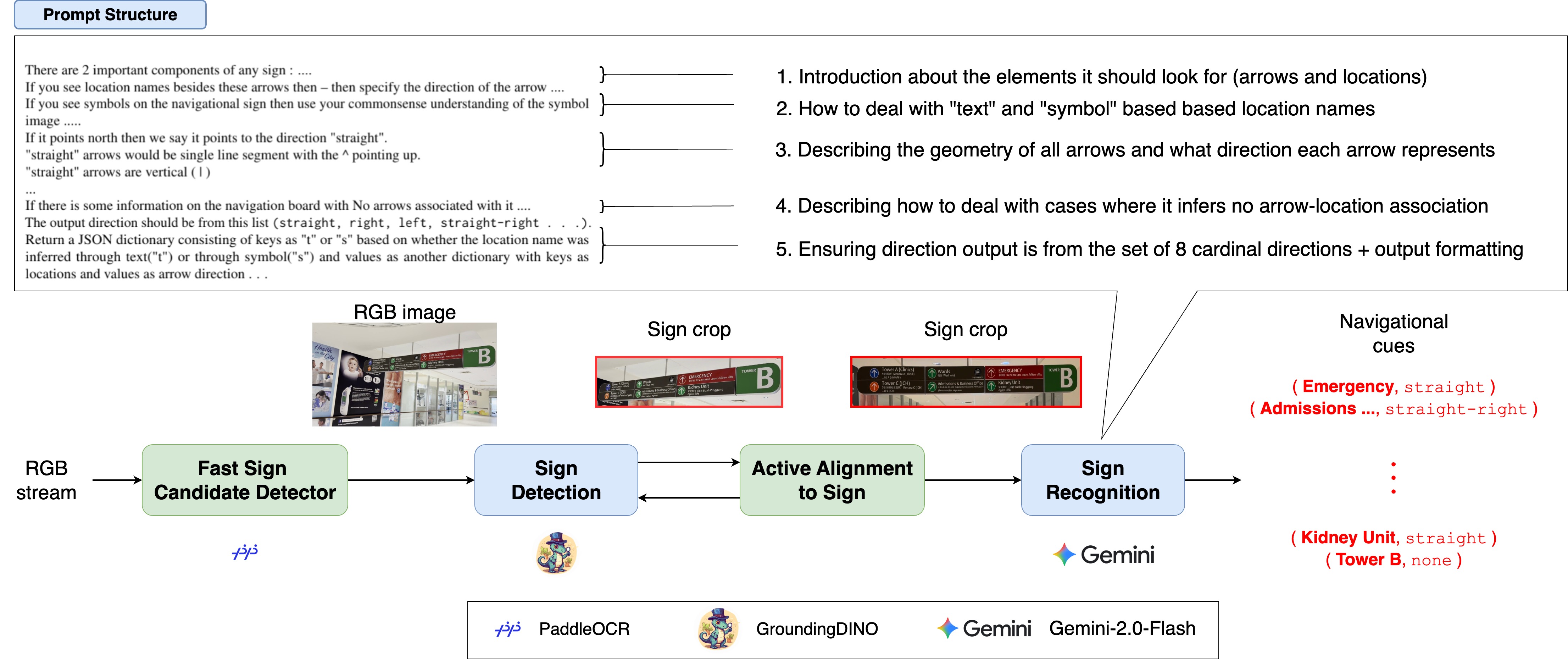}
  \caption{\textbf{Baseline.} Blue denotes core modules required for sign understanding. Green modules are to integrate sign understanding into robot systems: an efficient, approximate sign filter, and an alignment module which physically servos the robot to align with the sign's canonical viewing direction and optimize the view.}
  \label{fig:system}
\end{figure*}

\section{Dataset} 
\label{sec:data}

We curate a dataset to benchmark navigational sign understanding, and illustrate it in \figref{fig:dataset}. We consider two key dimensions of variation in this dataset: \emph{scene complexity} and \emph{sign complexity}.

% We curate a dataset on which to benchmark navigational sign understanding approaches, and visualise some samples from it in \figref{fig:dataset}. We consider two key dimensions of variation for in-the-wild navigational sign understanding: \emph{scene complexity} and \emph{sign complexity}.

%% Variation in scene appearance.
%% Occlusion
%% Distractors

\textbf{Scene complexity.} Environmental factors such as illumination, background textures, visual clutter, and occlusion affect perception of signs. Complexity also arises from non-navigational signs (e.g., regulatory signs) or sign-like objects (e.g., digital advertisements with arrows or location names) that can mislead sign detection.

% \textbf{Scene complexity.} This relates to environmental factors of variation that may affect both navigational sign detection and recognition. Variation in appearance and scene structure, including illumination, background textures, visual clutter and physical occlusions of the sign add to challenges in perceiving signs. In particular, non-navigational signs (\eg{} regulatory signs offering information unrelated to navigation), or objects that appear to be signs (\eg{} transient advertisements shown on screens that highlight a location name, or show arrows) add to the complexity of sign understanding in the wild.

%% Content: symbols, language, locational mixed, complex associations & multiple elements
%% Appearance: size and shape, illumination

\textbf{Sign complexity.} Navigational signs vary in appearance, shape, and structure. Such signs may contain multiple navigational cues of different types---locational (indicating a place) and directional (linking a place with a direction)---some of which may be presented in stylized or ambiguous ways, presenting difficulty in parsing these cues. For example, a single arrow may represent several destinations, requiring spatial or semantic reasoning to disambiguate. Signs also differ in formatting, symbols, and languages.

% \textbf{Sign complexity.} This relates to factors of variation in the structure and design of a navigational sign that may affect navigational sign recognition. Navigational signs vary greatly in terms of visual appearance, dimensions and shape, which can affect perception. More importantly, there is large variation in how the content on navigational signs is structured and presented. Signs may contain multiple navigational cues, which may further be of different types. In particular, locational cues (indicating a location) and directional cues (associating a location and a direction) may co-occur within a single navigational sign. Signs may be stylized in ways such that directional cues are not explicit and directly readable. \Eg{} signs where a single directional arrow is used to represent multiple locations, thus requiring some spatial and semantic reasoning to associate each location with a direction and output the directional cues. We also note that navigational signs may contain different languages and use a wide range of symbols.

\textbf{Dataset structure.} Our dataset reflects scene and sign complexity across diverse environments. It comprises two splits to evaluate sign detection and sign recognition respectively. The \textbf{sign detection} split contains 160 RGB images of scenes ranging across hospitals, malls, and campuses, each with human-annotated bounding boxes of navigational signs, $\signbbox_{i}^{\text{gt}}$. The \textbf{sign recognition} split includes 205 cropped signs covering various aspects of sign complexity, including different stylizations, content formats, and variation in the use of symbols. Each navigational sign crop is annotated with navigational cues as tuples $[(\signplace{}_1, \signdir{}_1),\dots, (\signplace{}_T, \signdir{}_T)]$, where $\signplace{}$ is a location label and $\signdir{}$ a direction category. We focus on English signs, noting that multilingual extensions are feasible with recent language models~\cite{du2020arxiv}.

\section{Metrics} 
\label{sec:metrics}

We describe metrics to benchmark the individual subtasks of navigational sign detection and recognition, and also for the overall task of navigational sign understanding.

\textbf{Detection metrics.} Detection is a specialized object detection task focusing on the category of signs that contain navigational information. To evaluate this, we employ standard object detection metrics, specifically COCO metrics~\cite{lin2014eccv} based on average precision (AP) and recall (AR).

\textbf{Recognition metrics.} Each sign in the dataset is annotated with ground-truth navigational cues, \ie{} $\setcues_n$ for sign $n$. Predictions from sign understanding are matched to $\setcues_n$, and we report both aggregated and per-sign metrics. \textit{Aggregated metrics} evaluate overall ability to parse cues, computed as precision and recall over all cues in the dataset. \textit{Per-sign metrics} assess whether a system can fully parse individual signs, measured as the success rate of exactly recovering all cues from a sign.

% For each sign in the dataset, the ground truth navigational cues $\{(\signplace{}_1, \signlabel{}_1, \signdir{}_1),\dots,(\signplace{}_t, \signlabel{}_t, \signdir{}_t)\}$ are annotated. We match the ground truth with parsed navigational cues from a sign understanding system, and compute both aggregated and per-sign metrics with the matches. \textit{Aggregated metrics} provide an indication of ability to parse any navigational cue, and are reported as precision and recall computed by combining all cues across all signs in the dataset. \textit{Per-sign metrics} provide an indication of the ability to accurately parse the content of a sign, and are computed as the success rate of accurately parsing all cues from a sign.

% Let $\setgt{}$ be a dataset of signs, where $s\in\setgt{}$ is the set of ground truth cues for a single sign. Let $\setpred{}$ be the predictions, where $p\in\setpred{}$ are predictions for a sign in $\setgt{}$. Given that the total number of matched cues across the dataset is $M_{\text{cue}}$, the aggregated metrics are:
% \begin{equation}
%     \text{Precision} = \frac{M_{\text{cue}}}{\sum\limits_{i=1}^{\lvert\setpred{}\rvert}|\setpred{}_i|},\text{  Recall} = \frac{M_{\text{cue}}}{\sum\limits_{j=1}^{\lvert\setgt{}\rvert}|\setgt{}_j|}
% \end{equation}
% The per-sign metrics are:
% \begin{equation}
%     \text{Success Rate} = \frac{\sum\limits_{i=1,\dots,|\setgt{}|}\mathbb{I}[s_i = p_i | s_i\in\setgt{}, p_i\in\setpred{}] }{|\setgt{}|}
% \end{equation}
% where $s_i = p_i$ if there is a one-to-one matching between ground-truth and predicted cues.

Formally, let $\setgt{}$ be the set of ground-truth cues over all signs, where $s_n \in \setgt{}$ are cues for sign $n$, and $\setpred{}$ the predictions, with $p_n \in \setpred{}$ the predictions corresponding to sign $n$. Let $M_{\text{cue}}$ denote the total number of matched cues across the dataset, then:
\begin{equation}
    \text{Precision} = \frac{M_{\text{cue}}}{\sum\limits_{i=1}^{\lvert\setpred{}\rvert}|\setpred{}_i|},\text{  Recall} = \frac{M_{\text{cue}}}{\sum\limits_{j=1}^{\lvert\setgt{}\rvert}|\setgt{}_j|}
\end{equation}
The per-sign metric is defined as:
\begin{equation}
    \text{Success Rate} = \frac{\sum\limits_{i=1}^{\lvert\setgt{}\rvert}\mathbb{I}[s_i = p_i] }{|\setgt{}|}
\end{equation}
where $s_i = p_i$ if there is a one-to-one match between predicted and ground-truth cues.

% A match between ground truth and parsed cues is made if $\signlabel{}_{t}^{\text{gt}} = \signlabel{}_{t}$, $\signdir{}_{t}^{\text{gt}} = \signdir{}_{t}$, and $\signplace{}_{t}^{\text{gt}}$ is equivalent to $\signplace{}_{t}$. We define several forms of equivalence to handle the fact that places may be specified via text or symbols. For places specified as text ($\signlabel{}_t = \texttt{text}$), we consider both strict equivalence as an exact string match of $\signplace{}_t^{\text{gt}}, \signplace{}_t$, and a relaxed equivalence only requiring that $\signplace{}_t$ is a substring of $\signplace{}_t^{\text{gt}}$. Since textual cues often refer to specific places or instances (\eg{} ``Tower B''), we emphasize that strict equivalence is most representative of performance on the task, although the relaxed criterion can offer insight into failure modes. For places specified as symbols ($\signlabel{}_t = \texttt{symbol}$), we define equivalence if the cosine similarity of the language embeddings of $\signplace{}_t^{\text{gt}}, \signplace{}_t$ exceed a specified threshold. Since text and symbols have different challenges and may be treated differently, we provide separate precision, recall and success rate metrics for cues containing text- and symbol-derived places, in addition to overall metrics that combine both.

A match requires $\signlabel_{t}^{\text{gt}} = \signlabel_t$, $\signdir_{t}^{\text{gt}} = \signdir_t$, and $\signplace_{t}^{\text{gt}}$ equivalent to $\signplace_t$. For locations indicated with text, we define strict equivalence as an exact string match and relaxed equivalence as substring containment. We consider strict matching to be most representative of actual task performance since textual cues often encode specific locations (e.g., “Tower B”). For locations indicated as symbols, we consider two symbols equivalent if the cosine similarity of their language embeddings exceeds a threshold. Because text and symbol cues pose distinct challenges, we report precision, recall, and success rate separately for text and symbol cases, in addition to overall metrics.

\textbf{Overall metrics.} To assess overall performance of navigational sign understanding, we define a sign to be \textit{understood} if it is both detected and if the extracted cues match the ground truth exactly. We further define \textbf{Precision$_{\texttt{sign}}$} as the fraction of perfectly parsed signs over the predicted set of human readable signs, and \textbf{Recall$_{\texttt{sign}}$} as fraction of perfectly parsed signs over the ground truth set of all human readable ground truth signs (obtained through \secref{sec:human-analysis}).

\section{Baseline} 
\label{sec:baseline}

We introduce a baseline for navigational sign understanding (\figref{fig:system}), that can be integrated on real robots for online operation. Our baseline relies on VLMs to detect navigational signs and extract navigational cues from them. For real robot integration, we add \textbf{(i)} an efficient filter to reduce calls to computationally expensive sign detection, \textbf{(ii)} a servoing step that aligns the robot (and sensor) with the sign head-on. 

Alignment is important for two reasons. Firstly, signs are designed to be viewed head-on, and the directions contained within are expressed with respect to this canonical viewing direction. For a robot to interpret a sign and take actions informed by it, the view angle must be accounted for. Secondly, we find empirically that VLMs' results are unreliable at oblique viewing angles.

% Say something about onboard/offboard compute

\subsection{Navigational Sign Understanding}

We use an open-set object detector to perform navigational sign detection. Specifically, we prompt GroundingDINO~\cite{liu2024grounding} to identify ``navigational signs'', returning 2D bounding boxes which we convert into image crops of each sign. The image crops are passed to a VLM that is prompted to parse text and symbols from a sign, and reason about the relationships between the entities on the sign. 

% \textbf{Sign legibility filtering.} As text on detected signs may be illegible (\eg{} due to distance, or wear) the system filters out signs with little readable text as these are unlikely to contain much useful semantic cues. To do so, we apply the Paddle-OCR model~\cite{du2020arxiv} to parse text from the sign crop and threshold based on the amount of detectable textual content.

% \textbf{View canonicalization.} Signs are often observed from rotated or skewed views in the wild, which we empirically observe to be more out-of-distribution for text recognition and foundation models, leading to higher error rates. View canonicalization aims to transform sign image crops to more in-distribution head-on, horizontally-aligned views before parsing. Our baseline attempts to orient the sign approximately horizontally. It computes the sign's major and minor axis, by applying PCA~\cite{pearson1901pmjs} to a mask of the sign from Grounded-SAM~\cite{ren2024arxiv}, then rotates the axis closest to the horizontal to align with the horizontal. We find that simply rotating the sign so that its text is horizontally aligned significantly improves parsing performance with foundation models. Addressing skewed viewpoints (\eg{} by computing a homography) is left to future work.

Given a sign crop, we query a VLM with the crop and a structured prompt to guide it in parsing the sign. Fig \ref{fig:system} highlights the structure of the prompt used. The prompt guides the VLM to \textbf{(i)} extract all location and place-related text from the image, \textbf{(ii)} extract all directional symbols from the image, \textbf{(iii)} associate locations and directions, \textbf{(iv)} output the associated directional cues as a list of tuples as described in \secref{sec:task}.

\subsection{Integrating into robot systems}

Online sign understanding should be efficient and onboard where possible, and also ensure physical alignment of the robot with respect to the sign for optimal viewing.

As current open-set object detectors run at low frame rates on embedded hardware, we add a lightweight filter that selects frames likely to contain candidate navigational signs. Based on the heuristic that navigational signs likely contain text, we run a fast OCR model (PaddleOCR~\cite{du2020arxiv}) to select only frames with text to run the detector on. If specific text strings of interest are known beforehand (\eg{} room names from a floor plan), we use fuzzy string matching to select frames containing these strings.

We aim to physically align the robot with the sign’s viewing direction and ensure a close-up view that ensures the entire sign is visible for recognition. Using camera intrinsics, we apply a monocular depth estimation model (Metric3Dv2~\cite{hu2024metric3d}) to generate an aligned depth map from RGB input. We obtain the centroid and surface normal of the sign, by estimating a best-fit plane using least squares on the sign's depth crop. From this, we compute a target position along the normal that both guarantees full visibility of the sign and maximizes its coverage in the image.

% \section{Active Perception} 
% \label{sec:activeperception}
% We provide an active perception pipeline for sign understanding to encourage the use of navigational signs in downstream tasks. Sign understanding in itself is a passive task, but exploiting navigational cues in the wild requires active measures to detect navigational sign candidates and position the robot for an optimal viewpoint, which increases the performance of the sign recognition. 

\section{Experimental Evaluation} 
\label{sec:exp}

\begin{figure*}[!t]
  \includegraphics[width=\linewidth]{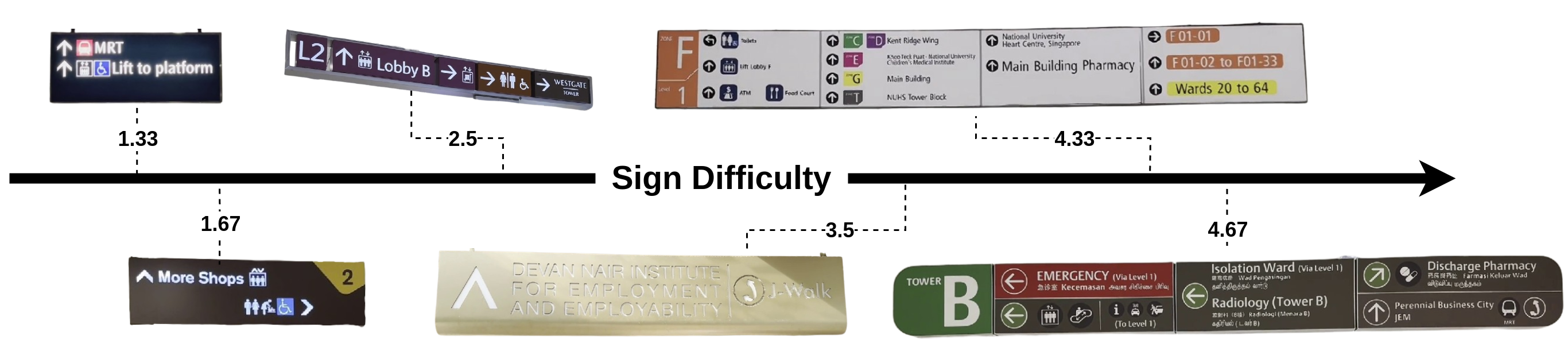}
  \caption{\textbf{Human perspective.} Sign recognition difficulty score (increasing left to right) for different signs ranked by the participants in the user study. Content overload, ambiguous place-arrow associations and nonstandard style contribute to the perceived difficulty of recognizing navigational signs.}
  \label{fig:diffranking}
\end{figure*}

%% Repeat the main focus/objective with one single(!) sentence starting with:

We consider the following questions:
\begin{enumerate}[label=\textbf{Q\arabic*.}]
    \item How accurately does our dataset capture the semantic information in navigational signs?
    \item How effective is our baseline at \textit{navigational sign detection}?
    \item How effective is our baseline at \textit{navigational sign recognition}?
    \item How effective is our baseline at the overall task of \textit{navigational sign understanding}?
    \item What aspects of scene or sign complexity are challenging for the baseline?
\end{enumerate}

\subsection{Sign Understanding from a Human Perspective}
\label{sec:human-analysis}
The task of sign understanding is efficiently performed by humans every day, and therefore we can safely use human performance as a safe heuristic to investigate \textbf{Q1}. Since unlike most learning approaches, human annotators can explain what factors contribute to the difficulty of sign detection and recognition -- we asked them to enumerate their reasons for the same. This analysis helped us illuminate the challenges inherent in the task.
% \textcolor{red}{What do we mean by this next line?}
% These task-specific challenges are not biased by our choice of a baseline.  

We conducted a study with ten participants. We tasked participants to rate the difficulty of detecting/recognizing signs and qualitatively describe factors affecting their ability to do so, based on fifteen signs sampled from the sign detection split. To verify correctness of ground-truth annotations for sign recognition, we also tasked participants to manually annotate six signs drawn the sign recognition split, deemed to be challenging (scored at difficulty $\geq 4$).

% For each sign we gave full image (from the sign detection split of our dataset) and the corresponding cropped sign (from sign recognition split of our dataset). While full images were used to understand the factors that affected the task of sign detection -- the cropped signs were used to used to evaluate the difficulty level and understand the factors that affected sign parsing. The participants were also asked to annotate the crops of last two signs, to verify the correctness of the ground truth annotations.

\figref{fig:diffranking} shows signs from the survey, with average difficulty score given by participants. We find that sign detection is most affected by \textbf{(a)} sign placement and \textbf{(b)} sign size. For sign recognition, participants highlight \textbf{(a)} ambiguous place-arrow associations, \textbf{(b)} content overload and \textbf{(c)} nonstandard stylizations of symbols/arrows as key challenges in parsing. To understand if signs perceived as difficult by participants are also challenging for our baseline, we evaluate our baseline on the set of signs annotated by participants. We see lower recognition success rate ($\sim15\%$) compared to the recognition performance in \secref{sec:exp_recog} ($\sim40\%)$, indicating that VLMs indeed struggle with such signs.

% We found that the factors affecting sign detection are \textbf{(a)} sign placement and \textbf{(b)} its size -- while the factors affecting sign parsing are \textbf{(a)} ambiguous place-arrow associations, \textbf{(b)} content overload and \textbf{(c)} nonstandard stylizing for symbols and/or arrows. A few samples from the sign parsing evaluation pool and their user-ranked sign parsing difficulty score are presented in \figref{fig:diffranking}.

% To have a better understanding about \textbf{Q1}\textemdash we evaluated our baseline on parsing a subset of signs that were rated at a difficulty $\ge$4 by human annotators. \textit{We observed a severe degradation in recognition success rate ($\sim 15\%$) as compared to the performance of our baseline's recognition module on the remaining sign recognition  split of our dataset ($\sim $~40\%)}. This suggests that the signs which were deemed difficult for humans were also difficult for the current VLMs. We discuss the reasons while discussing \textbf{Q5} in \secref{sec:discussion}.

We find that participants' annotations have a high degree of agreement with each other. The annotations proposed by the participants were compared against the ground truth annotation, and found to be matching in 95.5\% of the labels. This showcases a high level of consensus between the annotations and the ground truth, indicating that the ground truth annotations captures the semantic information encoded in the navigational signs. This also gave us the consensus on the notion of ``human readability'' of a sign. We further use this in analyzing \textbf{Q4}.

\begin{table}[h]
\centering
\caption{\textbf{Evaluating navigational sign detection:} We evaluate detection with AP and AR at varying IoU thresholds.}
\scriptsize
\begin{tabular}{@{} l c c c @{}}
\toprule
\textbf{Metric} & \textbf{MaxDets} & \textbf{GroundingDINO} & \textbf{Gemini-2.0-Flash}\\
\midrule
AP@[IoU=0.50] (all)        & 100 & 0.673  & 0.289\\
AP@[IoU=0.75] (all)        & 100 & 0.582 & 0.125 \\
AP@[IoU=0.25:0.75] (all)   & 100 & 0.657 & 0.288\\
\midrule
AR@[IoU=0.25:0.75] (all)   & 1   & 0.422  & 0.333\\
AR@[IoU=0.25:0.75] (all)   & 10  & 0.837 & 0.529\\
AR@[IoU=0.25:0.75] (all)   & 100 & \textbf{0.890} & 0.529\\
\bottomrule
\end{tabular}
\label{tab:sign_detection_all_results}
\end{table}

\begin{table}[h]
\centering
\caption{\textbf{Evaluating size based navigational sign detection} Evaluation of detection on the basis of navigational sign size.}
\scriptsize
\begin{tabular}{@{}lcccc@{}}
\toprule
\textbf{Metric} & \textbf{Size} & \textbf{MaxDets} & \textbf{GroundingDINO} & \textbf{Gemini-2.0-Flash}\\
\midrule
AP@[IoU=0.25:0.75] & S  & 100 & {0.105} & 0.003\\
AP@[IoU=0.25:0.75] & M & 100 & 0.190 & 0.039 \\
AP@[IoU=0.25:0.75] & L  & 100 & 0.774 & 0.385 \\
\midrule
AR@[IoU=0.25:0.75] & S  & 100 & 0.513  & 0.021\\
AR@[IoU=0.25:0.75] & M & 100 & 0.723 & 0.173\\
AR@[IoU=0.25:0.75] & L  & 100 & \textbf{0.949} & 0.637\\
\bottomrule
\end{tabular}
\label{tab:sign_detection_size_results}
\end{table}

  \begin{figure*}[!t]
    \centering
    \includegraphics[width=0.95\textwidth]{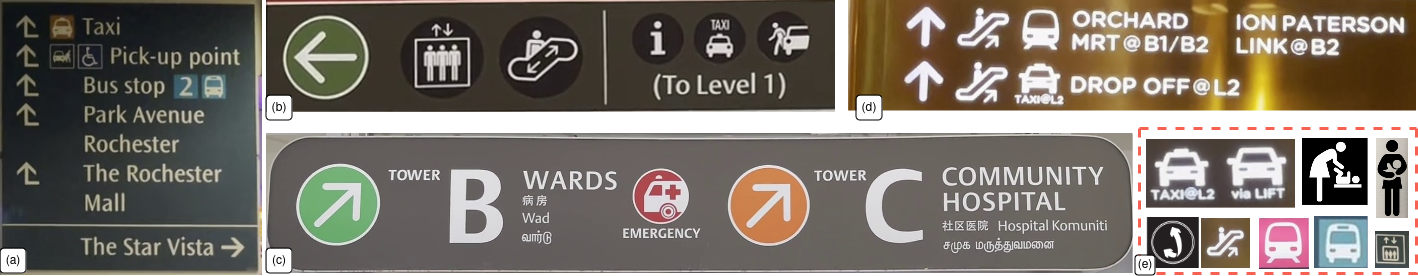}
    \caption{\textbf{Sign Recognition.} Some of the common types of failure cases and complexities observed in the task of sign recognition}
    \label{fig:discussion-figure}
\end{figure*}
\begin{figure}[!t]
  \includegraphics[width=\linewidth]{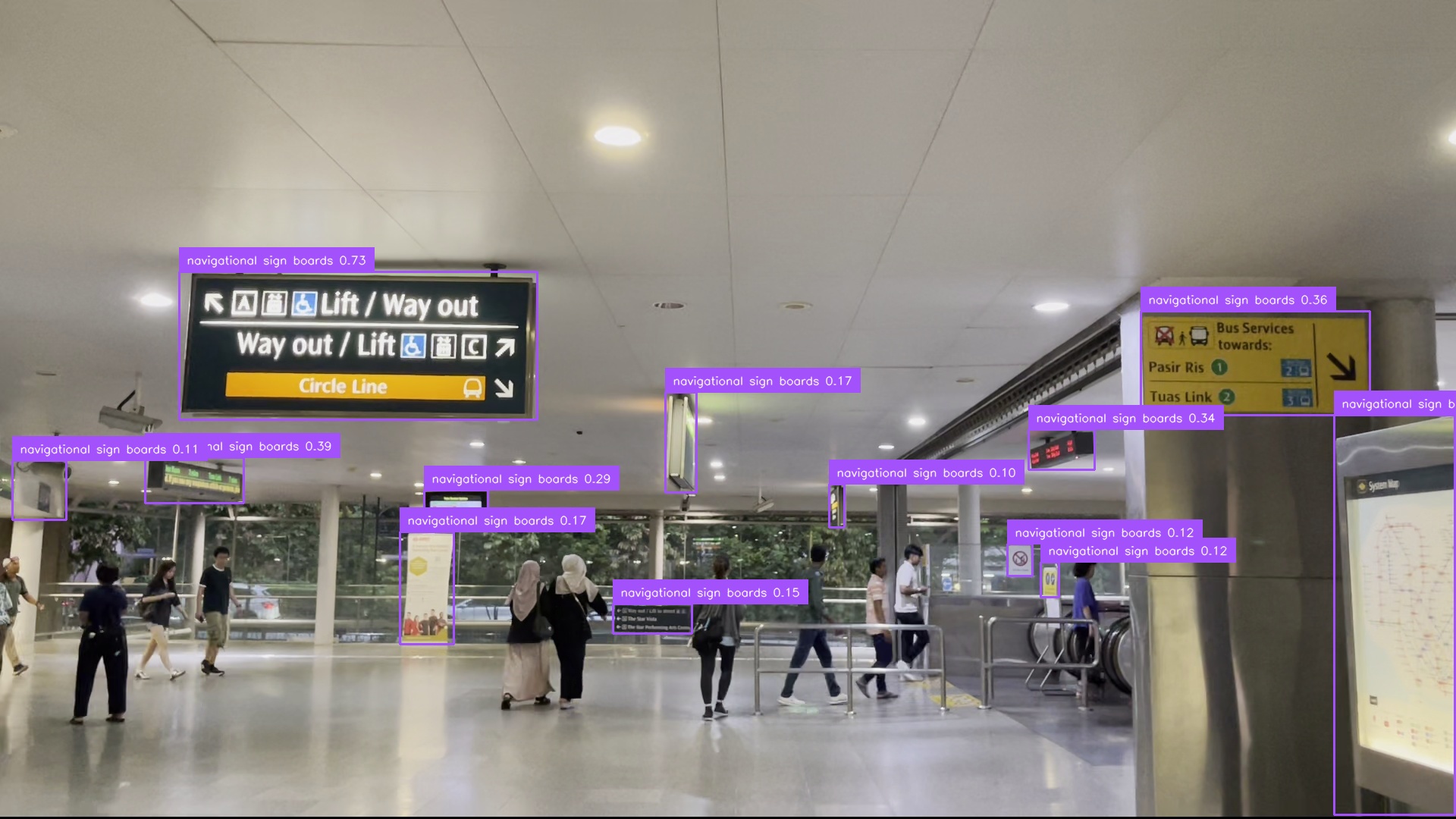}
  \caption{\textbf{Sign Detection.} An example annotation obtained by GroundingDINO. Here we observe that the confidence is higher for larger signs whereas the confidence is lower for the smaller, more distant signs. As compared to other model, the coherence in selecting signs was more evident in GroundingDINO.}
  \label{fig:dino-example}
\end{figure}

%% AP lower than AR -- failure modes
%% Large jump, semantic content
%% Specialised model like GroundingDINO outperforms gemini
\subsection{Navigational Sign Detection Performance}
\label{sec:exp_det}
% Here the input is an image from our evaluation dataset which could contain more than 1 navigational sign per image. \tabref{tab:sign_detection_all_results}, \tabref{tab:sign_detection_size_results} address \textbf{Q2} by quantifying detection performance with two different detector models. 

We address \textbf{Q2} by evaluating on the sign detection split of our dataset. Following the metrics in \secref{sec:metrics}, \tabref{tab:sign_detection_all_results} reports overall detection performance, while \tabref{tab:sign_detection_size_results} breaks down performance by sign size. Following COCO convention, we define signs in three sizes based on the bounding box area: Small (S) have $<32^2$ pixels, Medium (M) have $>32^2$ pixels and $<96^2$ pixels and Large (L) has $>96^2$ pixels. We compare detection performance of specialized object detectors (GroundingDINO~\cite{liu2024grounding}) with general VLMs (Gemini-2.0-Flash~\cite{team2023arxiv}).

Specialized object detectors outperform general VLMs on sign detection by a wide margin. Both models tend towards over-predicting signs, showing consistently higher recall than precision across all tests. Qualitatively, we find that models may have difficulty isolating \textit{navigational} signs from the variety of other signs present in human environments. (See \figref{fig:dino-example}) We also note that sign size strongly impacts the models' ability to positively identify a sign, with precision jumping significantly from medium to large signs for both models. We surmise that navigational signs may only be accurately distinguished from non-navigational signs on the basis of their information content, resulting in the models' performance improving dramatically for close-up views where the signs' content is clearly visible. We provide more detailed analysis and examples in \secref{sec:discussion}.

\subsection{Navigational Sign Recognition Performance}
\label{sec:exp_recog}

%% little difficulty with lexical understanding
%% higher difficulty with contextual
%% 

\begin{table}[tbp!]
\scriptsize
\setlength{\tabcolsep}{3pt}
\centering
\caption{\textbf{Ablations for navigational sign recognition:} Evaluation of baseline's performance in parsing text and symbols, using precision/recall. 
%Text parsing is evaluated for both cases in which \textbf{(i)} predictions and ground truth exactly match, \textbf{(ii)} predictions are a substring of the ground truth. 
}
\begin{tabular}{lccccccccc}
\toprule
\textbf{Model} & \multicolumn{3}{c}{\textbf{Precision}} & \multicolumn{3}{c}{\textbf{Recall}} \\[0.7ex]
\cmidrule(lr){2-4} \cmidrule(lr){5-7}
& \textbf{Txt (E)} & \textbf{Txt (S)} & \textbf{Sym} & \textbf{Txt (E)} & \textbf{Txt (S)} & \textbf{Sym} \\[0.5ex]
\midrule
GPT-4o & 0.79 & 0.80 & 0.82 & 0.59 & 0.60 & 0.70 \\
\rowcolor{gray!20}Gemini-2.0-Flash & 0.76 & 0.77 & 0.80 & 0.66 & 0.68 & 0.67 \\
\bottomrule
\end{tabular}
\label{tab:recognition_results}
\end{table}

\begin{table}[tbp!]
\caption{\textbf{Evaluating navigational sign recognition:} Evaluation of the overall success rate of recognition at a per-sign-level. A sign is defined as fully recognized (``overall'' columns) if \textbf{all} predicted cues (both symbol and text) match the ground truth cues. We also report the fraction of textual cues and symbolic cues correctly recognized at a per-sign level in the ``Txt/Sym Success Rate'' columns. 
%We consider both types of textual cues matching (\textbf{E}xact and \textbf{S}ub-string) for reporting overall accuracy.
}
% predicted and ground truth text cues to be matched \textbf{(i)} the text matches exactly (E), \textbf{(ii)} the predicted text cue is a substring of the ground truth cue (S). }
\scriptsize
    \begin{center}
        \begin{tabular}{cccccc}
            \toprule
            \textbf{Model} & \multicolumn{3}{c}{\textbf{Txt/Sym Success Rate}}& \multicolumn{2}{c}{\textbf{Success Rate}}\\
            \cmidrule(lr){2-4} \cmidrule(lr){5-6}
             & \textbf{Txt (E)} & \textbf{Txt (S)} & \textbf{Sym} & \textbf{Overall (E)} & \textbf{Overall (S)} \\
            \cmidrule(lr){2-4} \cmidrule(lr){5-6}
            {GPT-4o} & 44.6 & 44.6 & 63.3 & 39.3 & 39.3  \\
            % \textit{Sonnet-3.7} & &   \\
            \rowcolor{gray!20}{Gemini-2.0-Flash} & 47.3 & 47.9 & 65.8 & 41.6 & 42.2  \\
            \bottomrule
        \end{tabular}
    \label{tab:sign_recognition_results}
    \end{center}
\end{table}

We address \textbf{Q3} by evaluating on the sign recognition split of our dataset, using the recognition metrics (See \secref{sec:metrics}). This split comprises curated sign crops observed with approximately head-on orientation, with legible text and symbols. Following metrics in \secref{sec:metrics}, \tabref{tab:recognition_results} reports \textit{aggregated metrics} and \tabref{tab:sign_recognition_results} \textit{per-sign metrics}. For matching symbols, we employ CLIP-based cosine similarity. For matching text, we show results for both exact (E) and substring (S) approaches. If multiple predictions are substrings of a single ground-truth navigational cue, we consider only the match with the longest substring. We compare both GPT-4o~\cite{hurst2024arxiv} and Gemini-2.0 Flash~\cite{team2023arxiv} VLMs in our evaluations.

Strong performance on aggregated precision and recall indicates that existing VLMs can capably parse locational information (text and symbols) and perform the semantic reasoning needed to associate them with the right directions. Gemini-2.0-Flash generally outperforms GPT-4o on sign recognition, fully predicting signs accurately more often~(\tabref{tab:sign_recognition_results}). We found GPT-4o to be more conservative and infer only subsets of cues in the sign, leading to significantly lower recall than Gemini-2.0-Flash. Qualitatively, we find that VLMs can often generate lexically correct answers, but tend to struggle with parsing text that requires contextual understanding---\eg{} text spread across multiple lines and hence requiring analysis of semantic continuity to understand. However, inferring symbolic locations does not require such contextual awareness, and thus we observe a performance gap between inferring textual and symbolic cues at both \textit{aggregated} and \textit{per-sign level}.

% Furthermore, both the models had increase in performance when we chose sub-string matching instead of exact string match for associating the predicted textual locations with the ground truth textual locations. 

% Moreover, while both models were better at inferring location names depicted by symbols, we observed GPT-4o to be slightly better as compared to Gemini-2.0-Flash. We also observe GPT-4o to be more conservative and inferring only a subset of cues present in the sign, leading to slightly higher precision but notably lower recall than Gemini-2.0-Flash. 
% Through \tabref{tab:recognition_results} we observed higher precision for GPT-4o at a navigational cue level and observed it to perform worse as compared to Gemini-2.0-Flash at recall for text navigational cues. 
% This means GPT-4o predicted more false negatives for text navigational cues and less false negatives for symbolic navigational cues as compared to Gemini-2.0-Flash. This also suggests the prediction of fewer false positives by GPT-4o in general as compared to Gemini-2.0-Flash. 

% We note that significant room for improvement still  remains, and further highlight the difficulties of this task in \secref{sec:discussion}, including major challenges like accurately associating directions with locations in crowded and complex signs.

% \textcolor{red}{AA: i can calculate the total matches made by text (exact and substring separately) and also the symbols and the respective scores obtained. It would be helpful to get an idea of the error rates in arrow. let me know if required}

%%%%%%%%%%%%%%%%%%%%%%%%%%%%%%%%

%%%%%
\begin{table}[tbp]
\caption{\textbf{Ablations for navigational sign understanding:} Evaluation of different combinations of models for the combined task of detection and recognition. We report the precision at a sign level and recall at a sign level. The first model refers to the model that generates bounding boxes and the second model is used to parse those crops. We only allow matching of bounding boxes which have a minimum of 0.5 IoU with any of the ground truth bounding box}
\scriptsize
    \begin{center}
        \begin{tabular}{cccc}
            \toprule
            \multicolumn{2}{c}{\textbf{Models}} & & \\
            \cmidrule{1-2}
            \textbf{Detection} & \textbf{Recognition} & \textbf{Precision$_\texttt{sign}$} & \textbf{Recall$_\texttt{sign}$} \\
            \midrule
            Gemini-2.0-Flash & GPT-4o & {31.0} & {24.2} \\
            Gemini-2.0-Flash & Gemini-2.0-Flash & {39.2} & {30.3} \\
           \rowcolor{gray!20}{GroundingDINO} & Gemini-2.0-Flash & {39.3} & {39.0}  \\
           GroundingDINO & GPT-4o & {35.3} & {35.1}  \\
            % \textit{g-dino + Gemini-2.0-Flash} & {} & {} & {} & {} & {}  \\
            \bottomrule
        \end{tabular}
    \label{tab:acc_full-pipeline-results}
    \end{center}
\end{table}

\begin{figure*}[!t]
  \includegraphics[width=\linewidth]{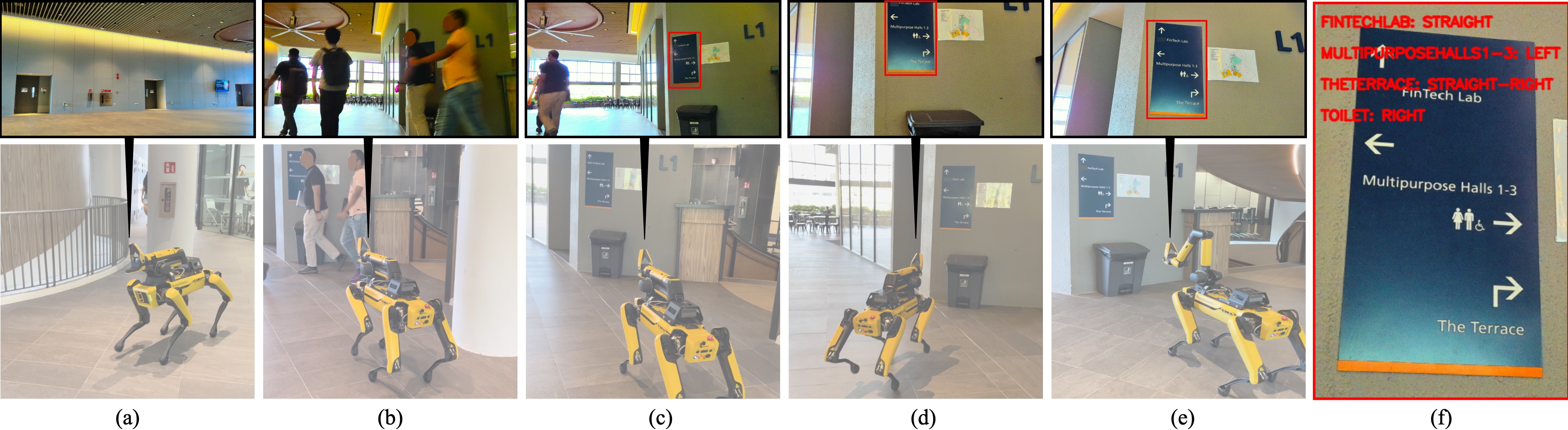}
  \caption{\textbf{Demo of plug-in module on Spot.} \textbf{(a, b)}: Spot explores the environment. \textbf{(c)}: Fast detector triggers sign detection, and overrides exploration upon positive detection. \textbf{(d, e)}: Plug-in module servos to face sign head-on, and arm makes adjustments to fully view sign. \textbf{(f)}: Successful parsing results obtained; Spot resumes exploration after.}
  \label{fig:demo_real_world}
\end{figure*}

%%%%%%%%%%%%%%%%%%%
\subsection{Navigational Sign Understanding Performance}

% Given that we have analyzed the performance of different models at the individual tasks of detection and recognition, we now combine these models into a modular setup to analyze their combined performance on the overall task of \textit{sign understanding}. As our problem statement is focused on parsing navigational cues on a sign level, we don't focus on zero shot VLM evaluations of full images. We prompt a model for bounding boxes and then match them to the maximum IoU ground truth bounding box (with a minimum threshold of 0.5 IoU). We then prompt these predicted bounding box crops to another model for parsing the information on a per sign level. Unlike the task of sign recognition where we fed in clean, readable and good view crops of signs \textemdash we don't constrain ourselves for this task. Here we feed in natural view full images but compare the parsed detections only to the human readable signs (as discussed in \secref{sec:human-analysis}. To quantitatively analyze \textbf{Q4}, we report the precision and recall at a \textit{sign level} (See \tabref{tab:acc_full-pipeline-results}). Our findings are consistent with our previous experiments where we found GroundingDINO to be rarely missing detecting navigational signs and Gemini-2.0-Flash to be more accurate in its sign parsing capabilities. The reduced performance as compared to \textbf{Q3} could be attributed to viewing angles, bounding box mis-detections and viewing distance. We also observed the bounding box detections of Gemini-2.0-Flash to be highly inconsistent as compared to GroundingDINO.

We evaluate the full baseline (\secref{sec:baseline}), which both detects signs and extracts navigational cues, to address \textbf{Q4}. Test inputs are drawn from the sign detection dataset split. We highlight that the dataset only annotates signs with legible text and symbols, and our metrics are computed based on this ground-truth. \tabref{tab:acc_full-pipeline-results} presents \textit{overall metrics} for variants of the baseline using different models for the sub-tasks of detection and recognition.  Consistent with above results, GroundingDINO continues to show strong performance on detecting navigational signs, while Gemini-2.0-Flash exhibits better performance in extracting navigational cues. We note that Gemini-2.0-Flash exhibits poorer recognition performance on signs cropped from the detection dataset split compared to evaluations on the curated recognition dataset split, likely owing to more variation in viewing angles and distance, and bounding box mis-detections. This highlights the need for good viewpoint selection for in-the-wild sign understanding.
% We also observed the bounding box detections of Gemini-2.0-Flash to be highly inconsistent as compared to GroundingDINO.

%%%%%%%%%%%%%%%%%%%%%%

\subsection{Discussion} 
\label{sec:discussion}

% In this section, we would like to discuss some of the common failure cases observed in our experiments. We believe this analysis would be of use to the research community to focus on addressing these common shortcomings in future upcoming works addressing the challenge of sign understanding. While we discussed about the common failure modes of sign detection task in previously, we now focus on common failure modes of sign understanding task. We divide the failures into the following sub-types. While these factors are a current limitation of foundation model pipelines, we believe this also makes the avenue of sign understanding a worthwhile problem statement to pursue.

We address \textbf{Q5} with a qualitative analysis of the common failure modes of navigational sign detection and recognition.

For \textit{navigational sign detection}, the main failure mode lies in semantically understanding whether the sign contains navigational cues. Empirically, we observe existing models like GroundingDINO perform well at identifying signs in general. However, distinguishing signs containing \textit{navigational cues} requires understanding of a sign's semantic content. Thus, detecting navigational signs is a challenge when the content of the sign is unclear (\eg{} due to distance) or when the content itself is semantically ambiguous (\eg{} large advertisement containing arrows and text).

For \textit{navigational sign recognition}, failure modes include:

\begin{itemize}
    %% hospital sign crop
    \item \textbf{Arrow association.} There is no guarantee of a one-to-one mapping between an arrow and location on a navigation sign. While it easy for us humans to infer such associations,
    models struggle to infer the intended direction from subtle context or commonsense cues~(\eg{}  \figref{fig:discussion-figure}(b) includes multiple symbols associated with a single arrow). 
    %and \figref{fig:discussion-figure}(c) include an ambagious association for the emergency label.
    % \textcolor{red}{can we use an example illustrating this failure? maybe to highlight why figuring out locational cues is no joke}
    
    % \item \textbf{Arrow Association.} Navigational signs don't necessarily have a directional arrow drawn for every location mentioned on them. In such cases the problem of deciding whether to associate it as a directional arrow vs a locational information boils down to our commonsense understanding and other subtle cues. In \figref{fig:discussion-figure}(b), would you associate "information-desk", "taxi", "car pickup" to "left" or "locational"(i.e. no arrow)? Similarly, in \figref{fig:discussion-figure}(c) even though there is no separate arrow for "emergency", we humans understand that "emergency" is towards "straight-right".

    %% J-walk, escalator, lift images
    % \item \textbf{Symbols With Arrows.} While we observed that these models have a good understanding of symbols \textemdash it was observed that they sometimes got confused in detecting directional arrows if an arrow was part of the symbol. In \figref{fig:discussion-figure}(e) we've shown the symbols of J-Walk, elevators, escalators that caused the most confusion. 
    \item \textbf{Symbols with arrows.} Confusion arises when arrows appear as part of a symbol (\eg{} elevator or escalator icons).
    
    %% Perennial business city JEM, Tower B wards
    % \item \textbf{Multi Line Location Names.} Location names on navigational signs are often written by considering the spatial constraints. While we humans have the contextual understanding of what's a locational name (owing to our familiarity or commonsense)\textemdash foundation models could still consider them as separate locations or merge them together. In \figref{fig:discussion-figure}(a) "Park Avenue Rochester" should be treated as 1 combined name. Similarly in \figref{fig:discussion-figure}(d) the correct tags should be "Orchard MRT@B1/B2" and "ION Paterson Link@B2" instead of breaking them at line level.
    \item \textbf{Multi-line names.} Spatial formatting leads models to split or merge location names incorrectly (\eg{} Fig. 5(a) “Park Avenue Rochester” parsed as multiple entities).

    % \item \textbf{Overcrowded Signs.} We observed that for big environments like hospitals, the navigational signs were composed of a lot of information. While we humans can  update our belief about our understanding of the sign by looking closely, foundation models yet don't have the understanding of when they need to look more closely. Thus for cases when the text/symbol was very small, it misclassified them. Overcrowded signs also increase the chances of mis-associations between text/symbol and directions. In \figref{fig:discussion-figure}(b) the direction for right part of the sign remains highly ambiguous.
    \item \textbf{Overcrowded signs.} Dense layouts with small text or symbols increase misclassification and incorrect text–direction associations.

    %% Taxi stand, car pickup, car park, taxi pickup, ....
    % \item \textbf{Ambiguous Symbols.} Decoding symbols is an ambiguous task. While sign designers try their best to convey the location name through symbols \textemdash it makes it more complicated sometimes. While some educated guess could be made by looking them closely or updating our belief by actually visiting the place\textemdash current models lack both of these abilities. In \figref{fig:discussion-figure}(e) we've shown few of the signs that were found to cause confusion in our baseline evaluations.(ex: "Bus Stop" and "MRT"). The symbols for "Car Park", "Car Pickup/Drop-off", "Taxi Stand", "Taxi Pickup/Drop-off" were found to be confusing even for our annotators. We observed similar failures to recognize these in our \textbf{Q3} and \textbf{Q4} analysis as well.

    \item \textbf{Ambiguous symbols.} Even for humans, certain symbols (\eg{} Fig. 5(e) car park vs. taxi pickup) are difficult to disambiguate, and models show similar confusions.

    %% Wheel chair access, bus stop 1, 2, 3
    %% Concierge sign
    % \item \textbf{Combinatorial Understanding.} Not all elements on a navigational sign correspond to a symbol or a text. There are cases where a text and symbol, or couple of symbols are to be read together. On other hand there are cases as well where some symbol elements don't add any information to the understanding of sign. The decision of where to combine elements and where to ignore is purely guided by our commonsense understanding of the navigational signs. For example: In \figref{fig:discussion-figure}(a) While the first entry is unaffected by the "Taxi" symbol, the second and third entry can't be properly understood without reading the symbols along with the text (i.e. "Wheelchair Accessible Car Pickup Point" and "Bus Stop No. 2"). Thus, decoding each element at a symbol and text level doesn't guarantee overall sign understanding. We believe that to make a general sign understanding system we need to move forward from parsing towards commonsense enabled interpretation. 
    \item \textbf{Combinatorial understanding.} Correct interpretation often requires reasoning over text–symbol combinations (e.g. Fig. 5(a) “Wheelchair Accessible Car Pickup Point”), which goes beyond element-wise parsing.
\end{itemize}

\section{Applications} 
\label{sec:conclusion}

We provide a plug-in module for online sign understanding on real robot hardware. We conclude by discussing potential downstream applications using semantic information from signs, which our plug-in sign understanding module enables.

\subsection{Sign Understanding on Real Robots}

% We demonstrate the plug-in sign understanding module from \secref{sec:baseline} on a Boston Dynamics Spot robot. \figref{fig:demo_real_world} shows the Spot exploring an environment, while runing the module concurrently. The module uses RGB input from the Spot's gripper camera, and runs fully onboard a Jetson Orin, with the Gemini VLM queried over 4G. A fast detector filters the high-rate RGB stream for images containing candidate signs at 10Hz, with selected images passed to the sign detector. Upon detecting a sign, the module overrides Spot’s navigation policy and iteratively servos toward the sign to optimize viewing angle and distance. The arm is actuated to make fine adjustments to the view when near the sign. The full servoing and sign understanding process completes in up to 20s.

We demonstrate the plug-in sign understanding module (\secref{sec:baseline}) on a Spot robot. Spot explores an environment (\figref{fig:demo_real_world}) while running the module onboard a Jetson Orin, using RGB input from the gripper camera and querying the Gemini-2.0-Flash VLM over 4G. A fast detector filters the RGB stream at 10Hz to identify candidate signs, which are then passed to the sign detector. Upon detecting a sign, the module overrides Spot’s navigation policy, and iteratively servos toward the sign to optimize angle and distance, with the arm making fine adjustments to optimize the view when close to the sign. The entire process completes within 20s.

\subsection{Downstream Applications}

Textual cues have been explored for localization~\cite{zimmerman2022iros, cui2021iros}, mapping~\cite{case2011icra}, and navigation~\cite{chen2025ral}, but prior work largely treated text as landmarks for map matching, overlooking the spatial information in directional signs. Talbot \etalcite{talbot2020tcds} highlighted their potential but offered no method to parse signs. Liang \etalcite{liang2020iros} obtained navigation policies from navigational signs, but made strong assumptions in parsing signs. In contrast, we propose a robust pipeline for sign understanding and a plug-in module for downstream applications.

Navigational signs contain information about places and their relative direction, which makes them useful for localizing in prebuilt maps that contain semantic and textual information. The place labels and associated directions are a constellation of abstract objects~\cite{ranganathan2007rss} that can be matched against the map to localize the robot. Similarly, for SLAM, navigational signs are good candidates for loop closures. Navigational signs are visually salient features, and their spatial semantic information can be used to verify the partially constructed map and the pose estimation. Navigational signs extend scene understanding beyond line-of-sight, and their compact semantic content can be easily integrated into representation such as scene graphs~\cite{gu2024icra, loo2025ijrr}, assisting with object-goal navigation and more efficient exploration.
The recent focus on topological and topo-semantic navigation~\cite{loo2025ijrr, shah2022arxiv}, as well as language goal specification~\cite{shah2023corl}, highlights the significance of navigational signs, due to their topo-semantic nature and their similarity to natural language. Further improvement in sign understanding can support navigational signs as a policy for navigation, increasing the impact and adoption of works such as Liang \etalcite{liang2020iros}.

% signs are spatial semantic cues. They can aid localization by being matched against the map. Similary, for SLAM, they are good candidates for loop closures, as the spatial semantic information can be matched against the map constructed so far.
% signs can be used as a policy - just following the place labels to a identically named goal
% signs can expand scene understanding by providng compact spatial semantic information about unseen places, which can be used for object goal navigation or effcient exploration
% Extracting navigational cues from signs opens new opportunities for autonomous agents. Signs can directly guide mapless navigation or navigation with incomplete maps in novel environments. They also provide a succinct topological description of key areas, useful for building semantic topological maps or aligning with existing maps for localization. 
%The recent focus on topological and topo-semantic navigation~\cite{shah2022arxiv, loo2025ijrr}, as well as language goal specification~\cite{shah2023corl}, highlights the significance of navigational signs, due to their topo-semantic nature and their similarity to natural language. 

Overall, navigational signs are a rich, underutilized source of semantic information that can greatly expand scene understanding capabilities of robots in human-oriented environments.

%\newpage
%%%%%%%%%%%%%%%%%%%%%%%%%%%%%%%%%%%%%%%%%%%%%%%%%%%%%%%%%%%%%%%%%%%%%%%%%%%%%%%%
%% Future work: Use only if applicable -- but if so, use the following
%% sentence to start:
% Despite these encouraging results, there is further space for improvements. 

%%%%%%%%%%%%%%%%%%%%%%%%%%%%%%%%%%%%%%%%%%%%%%%%%%%%%%%%%%%%%%%%%%%%%%%%%%%%%%%%
% Only if applicable
%\section*{Acknowledgments}
%We thank XXX for fruitful discussions and for \dots

\bibliographystyle{plain_abbrv}
% All new citations should go to new.bib. The file glorified.bib should go
% be the one from the ipb server. After paper or related work has been
% written merge the entries from new.bib to glorified.bib ON THE SERVER,
% replace the glorified.bib in this repository and empty the new.bib
\bibliography{glorified,new}

\end{document}